\documentclass[11pt]{article}

\usepackage[final]{acl}
\usepackage{graphicx}

\usepackage{times}
\usepackage{latexsym}

\usepackage[T1]{fontenc}

\usepackage{microtype}

\usepackage{inconsolata}

\usepackage{mathtools}
\usepackage{tabularx}

\title{Beyond What to Say: Proactive Utterance Timing \\with Inner Thoughts for Werewolf Agents}

\author{
  Yoshiki Tanaka, Shuya Uchiyama, Daichi Hamaguchi, Michimasa Inaba \\
  The University of Electro-Communications, Tokyo, Japan \\
  \texttt{\{y-tanaka, s-uchiyama, d-hamaguchi, m-inaba\}@uec.ac.jp}
}

\usepackage{booktabs}
\usepackage{placeins}


\begin{document}
\maketitle
\begin{abstract}
    Participating in a conversation involves not only deciding what to say but also whether to speak at all. This judgment carries even greater significance in environments like the Werewolf game, where participants' interests conflict, because both utterances and silence are subject to interpretation and assessment by others. Therefore, determining when to speak, as well as what to say, is crucial for an LLM agent playing the Werewolf game. In this study, we introduce a Werewolf AI agent developed for the Anytime Utterance Track of the AIWolfDial shared task. Our system builds on the existing Inner Thoughts framework, which determines utterance timing during conversations by generating internal thoughts and evaluating the value of each thought for expression at that moment. Since existing methods were designed for general dialogue, we have redesigned the system specifically for the competitive environment and incomplete information setting of Werewolf games. We call our adapted method Adversarial Inner Thoughts. Through experiments involving analysis of self-play logs and human evaluations, we analyze and discuss the effectiveness of our approach.

\end{abstract}

\section{Introduction}

Recent large language models (LLMs)~\cite{singh2026openaigpt5card,claude4-6} have demonstrated advanced conversational capabilities and are increasingly used as assistants that respond to user inputs. However, in human
communication, participating in a conversation involves not only deciding what to say but also choosing whether to speak at all. This choice is governed by conversational context and social norms,
and carries even greater significance in competitive dialogue environments.

The importance of choosing between speaking and remaining silent extends beyond everyday conversation to games that involve communication. A prominent example is the Werewolf game. The Werewolf game is an incomplete information game in which players are divided into a villager team and a werewolf team, each aiming to eliminate the opposing side without their roles being revealed. 
In this game, utterances containing contradictions may provide clues that allow others to infer one's true identity. Thus, utterances are constantly monitored and evaluated by other players, serving as evidence that influences suspicion and trust. At the same time, silence is also subject to interpretation. Refraining from speaking may be perceived as non-cooperation in the discussion and invite suspicion, while revealing too much information may make one a target of the werewolf team. In other words, the Werewolf game constitutes a dialogue environment in which both utterances and silence carry strategic significance.

The AIWolfDial shared task~\cite{kano-etal-2023-aiwolfdial,kano-etal-2024-aiwolfdial,kano-etal-2025-aiwolfdial} is an international competition that promotes the development of Werewolf AI agents. Its recently introduced Anytime Utterance Track allows each agent to freely decide when to speak, with a limit of four utterances per day. Details of the game rules and evaluation criteria are provided in Appendix~\ref{sec:task-details}.

Despite these efforts, LLM agent research targeting the Werewolf game has primarily focused on enhancing quality of utterance content and designing strategy~\cite{tanaka-etal-2024-enhancing,sato-etal-2024-implementation,qi-inaba-2024-enhancing,okada-ito-2025-towards}. For example, approaches have been proposed that employ prompt design to maintain consistency in utterances and actions, or that introduce personas to ensure diversity in expression. Consequently, the decision of when to speak and when to
remain silent has not been treated as a design variable. Meanwhile, research on utterance timing in dialogue has advanced in recent years~\cite{bhagtani-etal-2026-speak-or-stay,nama-etal-2026-when2speak,patel-etal-2026-discussllm,yang-etal-2026-proact,kimura-etal-2026-interruptible}, but most of it assumes settings in which only cooperation occurs, and does not consider environments where participants' interests conflict and silence functions as a strategy.

In this study, we formulate the Werewolf agent's utterance timing as a decision-making problem based on the evaluation of internal motivation. Specifically, we adapt the framework proposed by~\citet{liu-etal-2025-proactive} to the Werewolf game. In this framework, an agent generates internal thoughts behind the conversation and evaluates whether each thought is worth expressing at that moment, producing an utterance only when the evaluation exceeds a certain threshold. We redesign the evaluation criteria to suit the adversarial and incomplete information setting. Furthermore, since the Anytime Utterance Track imposes a limit on the number of utterances per day, speaking without discretion is costly, making this an environment where the effect of such control is particularly pronounced.

In summary, our contributions are as follows:
\begin{itemize}
    \item We propose Adversarial Inner Thoughts, an adaptation of the Inner Thoughts framework to multi-party dialogue in adversarial, incomplete-information games.
    \item We implement our proposed method for werewolf AI and analyze the impact of utterance timing control on agent behavior through game logs and human evaluations.
\end{itemize}

\section{Related Work}

\subsection{Agents for the Werewolf Game}

Since the Werewolf game requires reasoning under incomplete information, cooperation, persuasion, and deception, it
has served as a testbed for intelligent agents~\cite{toriumi-etal-2017-aiwolf,shibata-etal-2023-playing}. Recently, the AIWolfDial shared task~\cite{kano-etal-2023-aiwolfdial,kano-etal-2024-aiwolfdial,kano-etal-2025-aiwolfdial} has driven the development and
evaluation of agents that play the game through natural language conversation.

Advances in LLMs have enabled more flexible and sophisticated designs
for Werewolf agents. Early work examined whether LLMs can play the game
through free-form conversation without additional training~\cite{xu-etal-2023-exploring}. Later studies augmented LLMs with components that compensate for their limited strategic decision-making. An RL policy selects among LLM-generated action candidates~\cite{xu-etal-2024-language}, and an external module trained on human game records handles domain-specific logical inference~\cite{wu-etal-2024-enhance-reasoning}.
The AIWolfDial shared task has produced a range of such designs. To enhance the consistency and expressive quality of utterances, systems have summarized the dialogue history and conditioned generation on persona information~\cite{tanaka-etal-2024-enhancing}, or combined LLM-based generation with rule-based control~\cite{sato-etal-2024-implementation}.
To support reasoning and strategic play, other systems have explicitly represented the logical relations among players~\cite{watanabe-kano-2024-werewolf}, designed persuasion strategies grounded in situation analysis~\cite{qi-inaba-2024-enhancing}, and incorporated BDI-based reasoning
or manually designed game strategies into LLM agents~\cite{gondo-etal-2024-verification,okada-ito-2025-towards,harada-kano-2025-intent}. 

These studies share a common premise: a speaking opportunity is given, and the agent decides what to say within it. We instead give the agent the decision of whether to speak at all.

\begin{figure*}
  \centering
    \includegraphics[width=\linewidth]{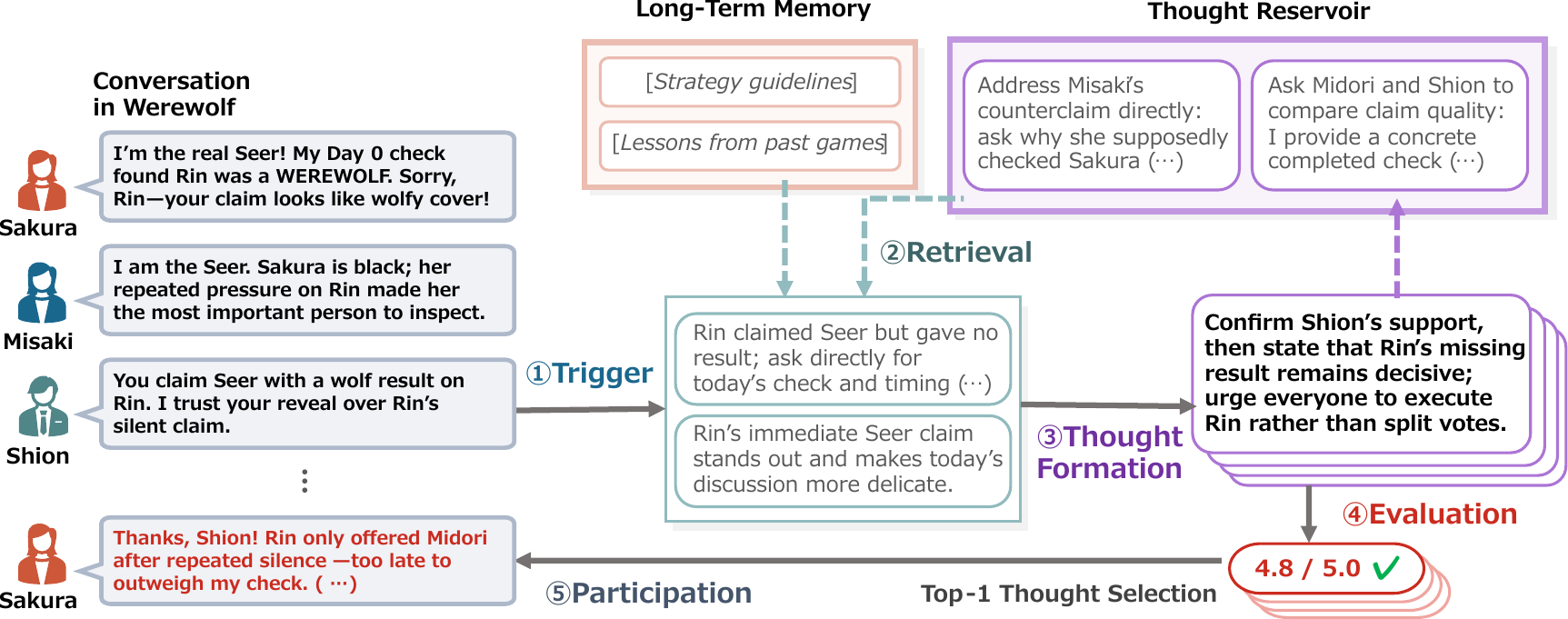}
    \caption{Overview of Adversarial Inner Thoughts, illustrated with a 5-player Werewolf game scenario. The agent (Sakura, Seer) receives a trigger from the ongoing conversation, retrieves relevant entries from long-term memory and the thought reservoir, generates a candidate inner thought via Thought Formation, and evaluates its score. Bracketed italics indicate the types of entries stored in long-term memory rather than actual entries.}
    \label{fig:proposed}
\end{figure*}

\subsection{Utterance Timing in Multi-party Dialogue}

Turn-taking is a fundamental mechanism by which participants coordinate who speaks and when~\cite{sacks-etal-1974-simplest-systematics}, and computational work has long studied how to predict it in spoken dialogue~\cite{ekstedt-skantze-2020-turngpt}. However, even strong LLMs are reported to know what to say but not when to say it~\cite{umair-etal-2024-large}. In text-based group chat, a message is complete when posted, so there is no turn end to detect, and the agent must decide at every moment whether to contribute at all.

This decision has been formulated explicitly as a choice between speaking and remaining silent. Recent work trains and evaluates agents that judge, at each point in an ongoing conversation, whether to contribute~\cite{bhagtani-etal-2026-speak-or-stay,nama-etal-2026-when2speak,patel-etal-2026-discussllm,yang-etal-2026-proact,kimura-etal-2026-interruptible}, indicating that appropriate silence has to be modeled or learned rather than expected to emerge from general language ability.  However, speakers also weigh whether an intended utterance fits the current position~\cite{jiang-etal-2023-response}, and the Inner Thoughts framework builds an agent around this view~\cite{liu-etal-2025-proactive}: it continuously generates private thoughts alongside the conversation and evaluates the intrinsic motivation to express each one, so that speaking and staying silent follow from what the agent is thinking rather than from the state of the conversation.

These settings are cooperative, and silence means that no useful contribution is available. In Werewolf, silence is itself strategic: an agent may withhold information that would expose its role, or delay an accusation and let another player's account settle. Two systems address timing here. Werewolf Arena lets agents bid for the floor~\cite{bailis-etal-2024-werewolf}, and \textit{Time to Talk} separates a scheduler deciding whether to speak from a generator producing the utterance~\cite{eckhaus-etal-2025-time}. In both, the decision is reduced to a single quantity computed without reference to any candidate utterance. \citet{eckhaus-etal-2025-time} suggest generating a candidate first and then judging whether to express it; we take this direction and ground the decision in the agent's own thoughts.

\section{Methodology}
In this study, we propose Adversarial Inner Thoughts, which
redesigns the Inner Thoughts framework by~\citet{liu-etal-2025-proactive} to adapt it to the adversarial, incomplete-information setting of the Werewolf game. Figure~\ref{fig:proposed} shows an overview of the proposed method.

\subsection{Original Framework}
In the original Inner Thoughts framework, an AI agent continuously generates internal thoughts in parallel to the ongoing conversation and decides when to speak based on its intrinsic motivation to express these thoughts. The framework operates through a five-step pipeline: (1) Trigger, (2) Retrieval, (3) Thought Formation, (4) Evaluation, and (5) Participation. This framework was
originally tested under cooperative, open-domain settings. We retain the pipeline structure and adapt each step to the Werewolf game, except for the Trigger step, which we adopt directly as its mechanism is equally applicable. In addition, the agent maintains a game history and a  conversation summary, which are referenced as shared context throughout the pipeline.

\subsection{Trigger}
Since the original Trigger mechanism of detecting new utterances and pauses is equally applicable to the Werewolf game, we adopt it directly.
Specifically, we define two types of triggers: \texttt{on\_new\_message}, which is activated when an agent receives a new utterance; and \texttt{on\_pause}, which prompts the agent to think when no participant has spoken for a certain period of time $t_{\mathrm{pause}}$.

\subsection{Retrieval}

Once a trigger is activated, the agent retrieves relevant
information to formulate its thoughts. Our framework maintains two types of memory for dynamic retrieval: long-term memory and a thought reservoir. 
\textbf{Long-term memory} stores entries describing the agent's general strategies, beliefs, and knowledge related to the game. Unlike the thought reservoir, which accumulates automatically during play, long-term memory entries are managed by predetermined logic: entries can be inserted or removed at specific game phases depending on the agent's role and the game state.
On the other hand, the \textbf{thought reservoir} accumulates inner thoughts generated by the agent during the game, each stored with metadata including a timestamp and the source utterance that prompted the thought.

The agent retrieves the top-$k$ entries most salient to the latest utterance. In the original framework, saliency is computed by directly comparing the thought text with the query. In our method, we instead use the source information of each entry: the memory text itself for long-term memory entries, and the associated source utterance for thoughts in the reservoir. This allows the agent to retrieve previous thoughts triggered by similar conversational events, enabling it to build on or refine its earlier reasoning when a comparable situation arises.
\begin{equation*}
  \begin{multlined}
  \mathrm{Saliency}(x_{\mathrm{src}},u)
  = \\\max\bigl(
      \mathrm{sim}(x_{\mathrm{src}},u), 
  \mathrm{sim}(x_{\mathrm{src}},u_{\mathrm{interp}})
    \bigr)
    \cdot w_x \cdot d_x
  \end{multlined}
\end{equation*}
where $u_{\mathrm{interp}}$ is an LLM-generated interpretation of $u$ that paraphrases the utterance to capture its underlying intent, mitigating cases where the raw utterance is too brief or ambiguous to match relevant memory entries.  $\mathrm{sim}$ denotes a similarity function, $w_x$ is a manually assigned weight for memory entry $x$, and $d_x$ is a temporal decay factor defined as $d_x = \lambda^{(t - \tau_x)}$, where $t$ is the current
time, $\tau_x$ is the timestamp of the memory, and $\lambda$ is a decay constant.

This retrieval serves two purposes: it encourages the agent to generate thoughts consistent with its earlier reasoning in similar situations, maintaining coherence across the game; and it provides an opportunity to refine previous thoughts that were insufficiently developed when a comparable situation arises again.

\begin{table*}[t!]
    \centering
    \small
    \begin{tabularx}{\linewidth}{lX}
        \toprule
        Criterion & Description \\
        \midrule
        (a) Contextual Relevance & Whether the thought relates to the current topic, game phase, recent utterances, and public information. \\
        (b) Naturalness & Whether expressing the thought fits naturally into the conversation flow without unnecessary repetition or abrupt topic shifts. \\
        (c) Temporal Appropriateness & Whether now is the right time to express this thought; thoughts requiring immediate expression are rated higher. \\
        (d) Information Utility & Whether expressing the thought organizes or provides information useful for reasoning. \\
        (e) Strategic Utility & Whether expressing the thought contributes to reasoning, persuasion, voting decisions, or the agent's team victory. \\
        (f) Risk Control & Whether expressing the thought avoids contradicting past statements, raising suspicion, leaking secret information, or unintentionally revealing strategy. \\
        \bottomrule
    \end{tabularx}
    \caption{Evaluation criteria for inner thoughts.}
    \label{tab:eval-criteria}
\end{table*}

\subsection{Thought Formation}
\label{sec:method-thought-formation}
In the Thought Formation step, the agent generates multiple candidate inner thoughts for expressing during their participation. An inner thought is not the utterance itself, but a concise utterance policy that describes a reaction to recent conversation, a topic to raise, or a dialogue action to take.

Following \citet{liu-etal-2025-proactive}, who were inspired by dual process theory~\citep{evans2008dual} and observed in their formative study that human participants' thoughts during multi-party conversation fall into reactive and deliberative patterns, inner thoughts are categorized into two types: System~1 and System~2.
System 1 generates immediate, reactive thoughts based primarily on the most recent utterances. System 2 takes both the conversation history and the game state as input to generate deliberative thoughts about the current situation. This distinction is particularly relevant in the Werewolf game, where an agent needs both immediate reactions to maintain natural conversational flow and deliberative reasoning to plan strategic behavior. The two systems generate thoughts independently via separate prompts. System 1 produces a single thought per trigger, whereas System 2 generates three thoughts from a single prompt to encourage diversity among candidates.

While the original framework targets general multi-party conversation, our method generates thoughts as concrete utterance policies that lead to decision-making and dialogue actions in the Werewolf game. System 2 in particular generates game-specific policies such as proposing a direction for discussion, requesting confirmation of specific information or claims, and performing role-dependent deception or strategic persuasion. For example, an agent assigned the Possessed role may generate a thought to falsely claim the Seer role in order to protect the Werewolf.

\subsection{Evaluation}
\label{sec:method-evaluation}
In the Evaluation step, each generated thought in Section~\ref{sec:method-thought-formation} is evaluated for the strength of its motivation to be expressed, using an LLM-as-a-Judge approach~\cite{zheng-etal-2023-judging,liu-etal-2023-g-eval}.
We follow the multi-criteria evaluation framework of~\citet{liu-etal-2025-proactive} and restructure the criteria to assess utterance appropriateness in the Werewolf game.
Specifically, drawing on the original framework's evaluation criteria, we define three general conversational criteria---Contextual Relevance, Naturalness, and Temporal Appropriateness---and three game-oriented criteria---Information Utility, Strategic Utility, and Risk Control~(Table~\ref{tab:eval-criteria}). 
On Day 0, where conversation is limited to greetings and preliminary discussion, only the first three criteria~(i.e., Contextual Relevance, Naturalness, and Temporal
Appropriateness) are used. From Day 1 onward, all six criteria are applied.

Rather than scoring all six criteria independently, the LLM considers the factors most relevant to the current situation, weighs the reasons for and against expressing the thought, and outputs a single motivation score from 1.0 to 5.0.

\subsection{Participation}
Based on the evaluation scores in Section~\ref{sec:method-evaluation} and the current turn situation, the agent decides whether to speak and generates an utterance from the selected thought. The agent first classifies the current turn situation into three types using an LLM: \textbf{open turn}, where no specific participant is expected to speak; \textbf{turn allocated to self}, where the agent is expected to respond; and \textbf{turn allocated to others}, where another participant is expected to respond.

Let $r^* = \max_i r_i$ denote the highest evaluation score among the System 2 thoughts, and let $\theta_{\mathrm{speak}}$ and $\theta_{\mathrm{interrupt}}$ ($\theta_{\mathrm{speak}} < \theta_{\mathrm{interrupt}}$) denote the speaking and interrupt thresholds, respectively. The agent's speaking decision is as follows:
\begin{itemize}
    \item \textbf{Turn allocated to self}: The agent always speaks using the highest-rated thought.
    \item \textbf{Open turn}: The agent speaks if $r^* \geq \theta_{\mathrm{speak}}$.
    \item \textbf{Turn allocated to others}: The agent speaks only if $r^* \geq \theta_{\mathrm{interrupt}}$, as the agent should intervene only when the motivation to express a thought is strong enough to justify interrupting another player's expected response.
\end{itemize}
If no System 2 thought meets the threshold, a System 1 thought may be selected with a turn-type-dependent probability, enabling short reactive responses and preventing excessive silence. If no thought is selected, the agent remains silent.

\subsection{Implementation Details for AIWolfDial 2026 Shared Task}

We use \texttt{gpt-5.6-luna} for all pipeline steps. The pause trigger fires after $t_{\mathrm{pause}} = 10$ seconds of silence. For embedding-based retrieval, we use \texttt{text-embedding-3-small} with $k = 3$.
The speaking and interrupt thresholds are set to $\theta_{\mathrm{speak}} = 4.0$ and $\theta_{\mathrm{interrupt}} = 4.3$, determined empirically.

In this shared task, each agent is assigned a character profile by the game server, which includes a name, age, gender, and personality description. Since the profile may not specify a speaking style, our agent generates a brief speaking style description and example utterances from the given personality at the start of each game, which are used throughout the pipeline to maintain consistent characterization. At the end of each day, the agent generates a summary of the conversation, which is referenced in subsequent steps to efficiently capture participants' claims and stances.

In addition to the Inner Thoughts pipeline, we incorporate rule-based behaviors for game-specific situations. Regardless of role, each agent reserves its last utterance of each day for declaring a vote target. The Seer reports divination results at the beginning of each day using a single prompt, bypassing the pipeline. For the Possessed role, the strategy guideline for a false Seer claim is inserted into long-term memory at the start of Day 1 with a high weight ($w = 10^6$), ensuring it is almost always retrieved and thus strongly influencing thought generation toward the intended deception. In the 5-player village, if the Possessed survives to Day 2, the remaining players necessarily consist of one Werewolf, one villager-team player, and the Possessed; in this situation, the agent calls for cooperation with the Werewolf to secure a majority vote, using a single prompt outside the pipeline. In this shared task, we use long-term memory only for role-specific strategy guidelines such as the false Seer claim described above, leaving the accumulation of cross-game experience for future work. The prompt templates used in each pipeline step are provided in Appendix~\ref{sec:prompt-templates}.

\section{Experiments}
To demonstrate that our Adversarial Inner Thoughts framework produces appropriate and proactive utterance behavior in Werewolf game scenarios, we conducted a human evaluation based on self-play logs in Japanese. We recruited 10 participants via a crowdsourcing platform\footnote{\url{https://crowdworks.jp}} to evaluate a total of 60 simulated game logs, with 30 games per condition.

\begin{table*}[t!]
  \centering
  \small
  \begin{tabularx}{\linewidth}{lX}
  \toprule
  \textbf{Dimension} & \textbf{Statement} \\
  \midrule
  Anthropomorphism     & The dialogue felt natural and resembled a conversation between humans. \\
  Coherence            & The dialogue had a natural and consistent flow of discussion throughout. \\
  Engagement           & Each player was well engaged in the discussion. \\
  Turn Appropriateness & The timing of each player's utterances and silences was contextually and logically appropriate. \\
  Initiative           & Each player proactively participated in the discussion when necessary. \\
  Adaptability         & Each player adapted well to the development of the discussion and others' utterances. \\
  \bottomrule
  \end{tabularx}
  \caption{Evaluation dimensions and statements used in the human evaluation.}
  \label{tab:measurements}
\end{table*}

\begin{figure*}[t!]
  \centering
    \includegraphics[width=\linewidth]{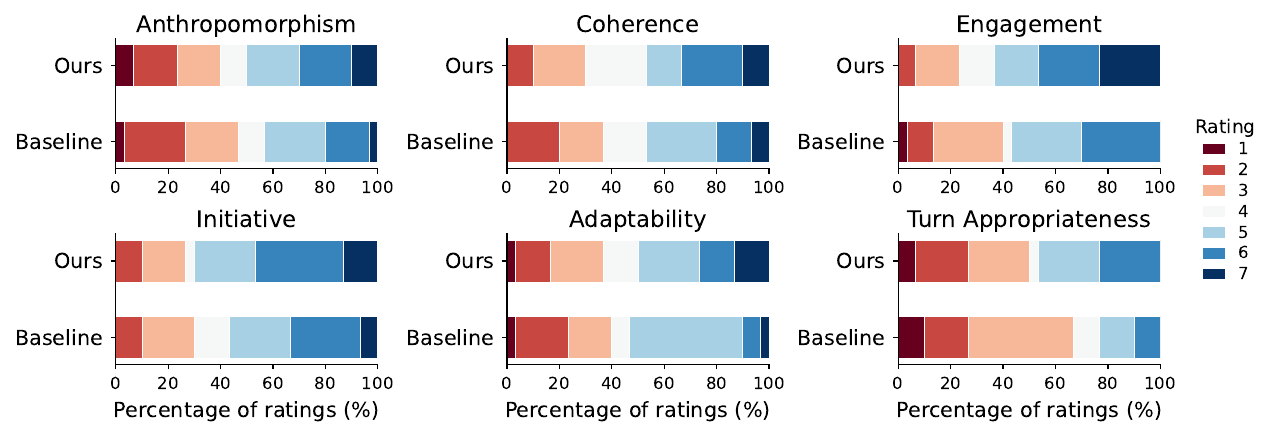}
    \caption{Distribution of 7-point Likert scale ratings for each evaluation dimension across the proposed and baseline conditions.}
    \label{fig:ratings_stacked}
\end{figure*}

\begin{figure*}[t!]
  \centering
    \includegraphics[width=\linewidth]{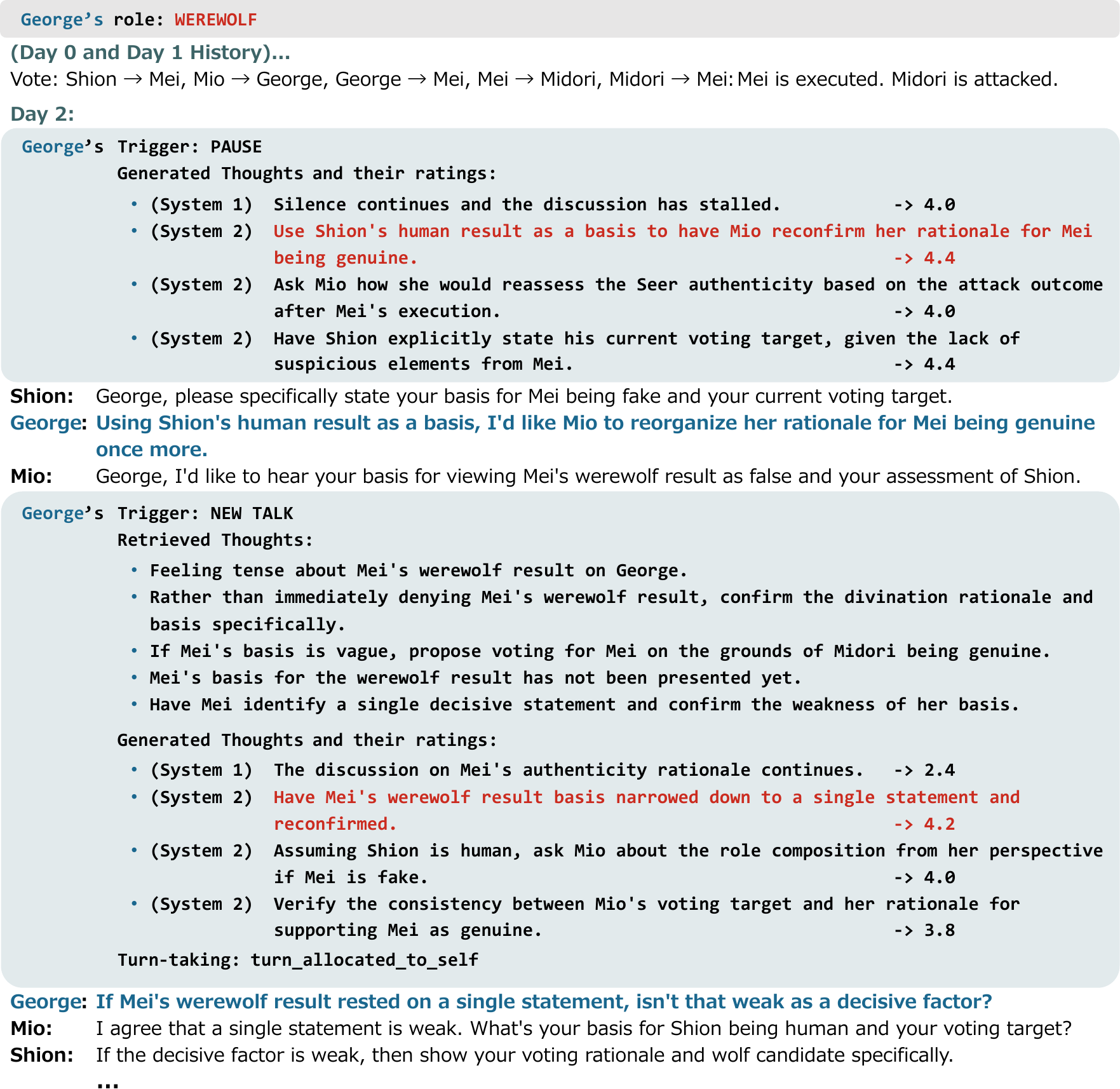}
    \caption{An example snippet of a self-play log and the proposed system's internal thought process on Day~2.}
    \label{fig:casestudy}
\end{figure*}

\subsection{Experimental Setup}
Following \citet{liu-etal-2025-proactive}, instead of displaying static game histories, we presented the logs as an animated chat interface that replays the game in real time, allowing participants to perceive utterance timing more accurately. The interface also displayed in-game events such as voting, execution, and attack results.

For the self-play generation, we applied the following controls. To ensure that the evaluation focused on game-relevant dialogue from Day~1 onward, each agent was restricted to a single utterance on Day~0. To control for game length, only games completed within eight minutes were included. In addition, since all agents share the same pause threshold $t_{\mathrm{pause}}$, they would initiate speech simultaneously when the threshold is reached. To avoid this, we added random noise sampled from $\mathcal{N}(0, 1)$ to each agent's trigger timing.

\subsection{Baseline System}
To verify that the improvement in dialogue quality stems from our thought generation and evaluation components rather than the trigger-based turn detection alone, we define a baseline system. 
The baseline is derived from our proposed system by retaining only the trigger-based turn detection, which classifies the current turn situation into one of three types: open turn, turn allocated to self, and turn allocated to others. As in our system, the agent always speaks in a turn allocated to self. Unlike our system, however, the baseline always remains silent in a turn allocated to others. In an open turn, the agent speaks with a probability of 0.3. Since utterance timing in this design depends solely on the current turn situation, all agents may simultaneously refrain from speaking, causing the conversation to stall completely. To address this, if a turn allocated to others occurs twice in a row, the subsequent turn is treated as an open turn to ensure speaking opportunities. For a fair comparison, the proposed system adopts the same fallback mechanism.

\subsection{Evaluation Procedure}
Prior to the evaluation, participants were provided with a manual explaining the rules of the five-player Werewolf game as well as the purpose and procedure of the study. At the start of the evaluation, participants answered a single question about their prior experience with the Werewolf game. They then completed a comprehension check consisting of seven questions about the game rules and roles to confirm their understanding of the game flow. Participants were required to answer all questions correctly; those who answered any question incorrectly were asked to retake the test until they passed.\par

Participants then watched six game logs in total. They were informed that all dialogues were generated by AI agents. The first three games were from one condition and the remaining three from the other, with the order of conditions counterbalanced across participants.
After watching each game, participants evaluated the dialogue quality on six dimensions. After watching all six games, participants completed a comparative evaluation between the two conditions.

\subsection{Measurements}
After watching each game, participants evaluated the quality of the dialogue rather than the agents' in-game strategies or tactics. They rated their agreement with six statements in Table~\ref{tab:measurements} on a 1--7 Likert scale ranging from
strongly disagree to strongly agree. Participants were also asked to describe any specific parts where utterance or silence timing felt unnatural in an open-ended response. After evaluating all games, participants selected which condition felt more natural and human-like from three options---the first half, tie, or the second half---and provided an explanation with concrete examples.

\subsection{Results}

Figure~\ref{fig:ratings_stacked} shows the distribution of ratings for
each dimension. Our system received higher mean scores than the baseline across all six dimensions (Anthropomorphism: 4.2 vs.\ 3.9; Coherence: 4.5 vs.\ 4.2; Engagement: 5.0 vs.\ 4.3; Turn Appropriateness: 3.9 vs.\ 3.3; Initiative: 4.9 vs.\ 4.6; Adaptability: 4.3 vs.\ 4.0). To account for the repeated-measures design, we aggregated the ratings per participant and compared the two conditions using the Wilcoxon signed-rank test. None of the differences reached statistical significance, with Engagement showing the smallest p-value ($p = .121$). In the comparative evaluation, four participants preferred the baseline, three preferred the proposed system, and three rated them as comparable.\par

The open-ended responses revealed qualitative differences between the two conditions. The most frequent criticism of the proposed system concerned non-responses to directed questions. One participant noted: ``\textit{Many players did not respond to questions from others, even when it was their chance to clear suspicion.}'' This is likely due to the higher latency of our pipeline, which requires five LLM API calls per trigger (retrieval, thought formation, evaluation, turn-taking, and response generation), compared to two in the baseline. The resulting delay may cause the agent to miss the appropriate timing for responding to a directed question. A possible improvement would be to introduce a streamlined response path for directly addressed turns, reducing the number of processing steps when the agent is explicitly named or when an immediate response is expected.
On the other hand, the baseline system was more frequently criticized for prolonged silences and inconsistent utterance pacing. One participant described the silences as ``\textit{unnaturally long, even considering deliberation time.}'' This suggests that its turn-taking mechanism, which relies solely on turn situation classification, struggles to produce well-timed utterances.\par

Interestingly, several participants also noted that the proposed system produced more substantive arguments supported by concrete reasoning. In the comparative evaluation, one participant commented that ``each player's claims were well supported by concrete reasoning, and the discussion felt more heated,'' while another noted that ``the utterances were aligned with others' opinions and the overall flow of the discussion, making it a substantive debate.'' This may be because the baseline only determines whether to speak based on who should take the turn, whereas the proposed system additionally evaluates whether each thought is worth expressing at that moment. Both conditions shared common issues such as repetitive expressions and translation-like phrasing, which are attributable to the underlying language model rather than the utterance timing mechanism.

\subsection{Case Study}
Figure~\ref{fig:casestudy} shows a Day~2 excerpt where three players remain: Shion and Mio (both Villagers) and George (Werewolf), who was declared werewolf by Mei, a counter-claiming Seer, on Day~1.

Once George's pause trigger fires, our system generates four candidate thoughts. It then selects ``Use Shion's human result as a basis to have Mio reconfirm her rationale for Mei being genuine'', generating an utterance that redirects the discussion toward the Seer's credibility. 
Note that Shion's utterance appears just 0.2 seconds earlier in the log due to asynchronous generation; George's response was generated independently upon detecting the silence. Immediately after, Mio directly addresses George, and the turn-taking classifier classifies this as \texttt{turn\_allocated\_to\_self}. The system then selects a thought that challenges the basis of Mei's werewolf result
(rated 4.2), producing a concrete rebuttal.

This example illustrates three key properties of our approach: the system treats silence not as passive waiting but as a trigger to initiate proactive speech; it evaluates the strategic value of each thought rather than simply filling the silence; and it adapts its response when directly addressed, shifting from a proactive stance to a context-appropriate reply.

\section{Conclusion}
In this paper, we adapted the Inner Thoughts framework to the adversarial, incomplete-information setting of the Werewolf game. Human evaluation showed that the proposed system numerically outperformed the baseline across all dimensions, though a trade-off between utterance quality and response timeliness was observed. Future work includes reducing pipeline latency and exploring the use of long-term memory to retain strategic experiences across games.

\section*{Limitations}
This study has several limitations. First, our evaluation is limited in scale: we collected ratings for 60 games from 10 participants, with each game rated by a single participant. This prevents us from measuring inter-annotator agreement and likely contributes to the lack of statistical
significance despite consistent differences in mean scores. A larger-scale evaluation with multiple ratings per game would allow us to draw firmer conclusions and to conduct finer-grained analyses, such as examining how evaluator familiarity with the Werewolf game affects ratings.

Second, the evaluation was conducted exclusively in Japanese with the five-player village configuration, and all games were self-play matches in which every agent used the same system. In the AIWolfDial shared task, agents compete against those developed by
other teams, and whether our approach functions equally well in such heterogeneous settings requires further investigation.

Another limitation is that our proposed pipeline affects not only utterance timing but also utterance content through the generation and evaluation of multiple candidate thoughts. Therefore, the current comparison does not isolate the effect of utterance timing itself. Future work should explore appropriate evaluation designs for disentangling the effects of utterance timing and content. For example, we could control dialogue content while manipulating only the timing of utterances.

Finally, the speaking and interrupt thresholds were determined empirically rather than through systematic search. Since the shared task assigns each agent a character profile, these thresholds could also be adapted to reflect individual personalities, for example by making talkative characters more prone to interrupting. We leave this direction to future work.


\bibliography{custom}

@inproceedings{liu-etal-2025-proactive,
    author = {Liu, Xingyu Bruce and Fang, Shitao and Shi, Weiyan and Wu, Chien-Sheng and Igarashi, Takeo and Chen, Xiang 'Anthony'},
    title = {Proactive Conversational Agents with Inner Thoughts},
    year = {2025},
    isbn = {9798400713941},
    publisher = {Association for Computing Machinery},
    address = {New York, NY, USA},
    url = {https://doi.org/10.1145/3706598.3713760},
    doi = {10.1145/3706598.3713760},
    booktitle = {Proceedings of the 2025 CHI Conference on Human Factors in Computing Systems},
    articleno = {184},
    numpages = {19},
    location = {
    },
    series = {CHI '25}
}

@inproceedings{kano-etal-2023-aiwolfdial,
    title = "{AIW}olf{D}ial 2023: Summary of Natural Language Division of 5th International {AIW}olf Contest",
    author = "Kano, Yoshinobu  and
      Watanabe, Neo  and
      Kagaminuma, Kaito  and
      Aranha, Claus  and
      Lee, Jaewon  and
      Hauer, Benedek  and
      Shibata, Hisaichi  and
      Miki, Soichiro  and
      Nakamura, Yuta  and
      Okubo, Takuya  and
      Shigemura, Soga  and
      Ito, Rei  and
      Takashima, Kazuki  and
      Fukuda, Tomoki  and
      Wakutani, Masahiro  and
      Hatanaka, Tomoya  and
      Uchida, Mami  and
      Abe, Mikio  and
      Mikami, Akihiro  and
      Otsuki, Takashi  and
      Qi, Zhiyang  and
      Harada, Kei  and
      Inaba, Michimasa  and
      Katagami, Daisuke  and
      Osawa, Hirotaka  and
      Toriumi, Fujio",
    editor = "Mille, Simon",
    booktitle = "Proceedings of the 16th International Natural Language Generation Conference: Generation Challenges",
    month = sep,
    year = "2023",
    address = "Prague, Czechia",
    publisher = "Association for Computational Linguistics",
    url = "https://aclanthology.org/2023.inlg-genchal.13/",
    pages = "84--100"
}

@inproceedings{kano-etal-2024-aiwolfdial,
    title = "{AIW}olf{D}ial 2024: Summary of Natural Language Division of 6th International {AIW}olf Contest",
    author = "Kano, Yoshinobu  and
      Sahashi, Yuto  and
      Watanabe, Neo  and
      Kagaminuma, Kaito  and
      Aranha, Claus  and
      Katagami, Daisuke  and
      Harada, Kei  and
      Inaba, Michimasa  and
      Ito, Takeshi  and
      Osawa, Hirotaka  and
      Otsuki, Takashi  and
      Toriumi, Fujio",
    editor = "Kano, Yoshinobu",
    booktitle = "Proceedings of the 2nd International AIWolfDial Workshop",
    month = sep,
    year = "2024",
    address = "Tokyo, Japan",
    publisher = "Association for Computational Linguistics",
    url = "https://aclanthology.org/2024.aiwolfdial-1.1/",
    doi = "10.18653/v1/2024.aiwolfdial-1.1",
    pages = "1--12"
}

@inproceedings{kano-etal-2025-aiwolfdial,
    title = "{AIW}olf{D}ial 2025: Summary of Natural Language Division of 7th International {AIW}olf Contest",
    author = "Kano, Yoshinobu  and
      Watanabe, Neo  and
      Harada, Yuya  and
      Sahashi, Yuto  and
      Aranha, Claus  and
      Katagami, Daisuke  and
      Harada, Kei  and
      Inaba, Michimasa  and
      Ito, Takeshi  and
      Osawa, Hirotaka  and
      Otsuki, Takashi  and
      Toriumi, Fujio",
    editor = "Kano, Yoshinobu",
    booktitle = "Proceedings of the 3rd International AIWolfDial Workshop",
    month = oct,
    year = "2025",
    address = "Hanoi, Vietnam",
    publisher = "Association for Computational Linguistics",
    url = "https://aclanthology.org/2025.aiwolfdial-1.1/",
    pages = "1--13"
}

@inproceedings{tanaka-etal-2024-enhancing,
    title = "Enhancing Consistency of {Werewolf} {AI} through Dialogue Summarization and Persona Information",
    author = "Tanaka, Yoshiki  and
      Kaneko, Takumasa  and
      Onozeki, Hiroki  and
      Ezure, Natsumi  and
      Uehara, Ryuichi  and
      Qi, Zhiyang  and
      Higuchi, Tomoya  and
      Asahara, Ryutaro  and
      Inaba, Michimasa",
    editor = "Kano, Yoshinobu",
    booktitle = "Proceedings of the 2nd International AIWolfDial Workshop",
    month = sep,
    year = "2024",
    address = "Tokyo, Japan",
    publisher = "Association for Computational Linguistics",
    url = "https://aclanthology.org/2024.aiwolfdial-1.6/",
    doi = "10.18653/v1/2024.aiwolfdial-1.6",
    pages = "48--57"
}

@inproceedings{watanabe-kano-2024-werewolf,
    title = "{Werewolf} Game Agent by Generative {AI} Incorporating Logical Information Between Players",
    author = "Watanabe, Neo  and
      Kano, Yoshinobu",
    editor = "Kano, Yoshinobu",
    booktitle = "Proceedings of the 2nd International AIWolfDial Workshop",
    month = sep,
    year = "2024",
    address = "Tokyo, Japan",
    publisher = "Association for Computational Linguistics",
    url = "https://aclanthology.org/2024.aiwolfdial-1.3/",
    doi = "10.18653/v1/2024.aiwolfdial-1.3",
    pages = "21--29"
}

@inproceedings{sato-etal-2024-implementation,
    title = "An Implementation of {Werewolf} Agent That does not Truly Trust {LLM}s",
    author = "Sato, Takehiro  and
      Ozaki, Shintaro  and
      Yokoyama, Daisaku",
    editor = "Kano, Yoshinobu",
    booktitle = "Proceedings of the 2nd International AIWolfDial Workshop",
    month = sep,
    year = "2024",
    address = "Tokyo, Japan",
    publisher = "Association for Computational Linguistics",
    url = "https://aclanthology.org/2024.aiwolfdial-1.7/",
    doi = "10.18653/v1/2024.aiwolfdial-1.7",
    pages = "58--67"
}

@inproceedings{qi-inaba-2024-enhancing,
    title = "Enhancing Dialogue Generation in {Werewolf} Game Through Situation Analysis and Persuasion Strategies",
    author = "Qi, Zhiyang  and
      Inaba, Michimasa",
    editor = "Kano, Yoshinobu",
    booktitle = "Proceedings of the 2nd International AIWolfDial Workshop",
    month = sep,
    year = "2024",
    address = "Tokyo, Japan",
    publisher = "Association for Computational Linguistics",
    url = "https://aclanthology.org/2024.aiwolfdial-1.4/",
    doi = "10.18653/v1/2024.aiwolfdial-1.4",
    pages = "30--39"
}

@inproceedings{harada-kano-2025-intent,
    title = "Intent-driven {AIW}olf Agents with Hierarchical {BDI} Model and Personality",
    author = "Harada, Yuya  and
      Kano, Yoshinobu",
    editor = "Kano, Yoshinobu",
    booktitle = "Proceedings of the 3rd International AIWolfDial Workshop",
    month = oct,
    year = "2025",
    address = "Hanoi, Vietnam",
    publisher = "Association for Computational Linguistics",
    url = "https://aclanthology.org/2025.aiwolfdial-1.3/",
    pages = "22--36"
}

@inproceedings{okada-ito-2025-towards,
    title = "Towards a Strategic {Werewolf} {AI} Based on Expert Strategies in Five-Player {Werewolf}",
    author = "Okada, Takuma  and
      Ito, Takeshi",
    editor = "Kano, Yoshinobu",
    booktitle = "Proceedings of the 3rd International AIWolfDial Workshop",
    month = oct,
    year = "2025",
    address = "Hanoi, Vietnam",
    publisher = "Association for Computational Linguistics",
    url = "https://aclanthology.org/2025.aiwolfdial-1.2/",
    pages = "14--21"
}

@inproceedings{gondo-etal-2024-verification,
    title = "Verification of Reasoning Ability using {BDI} Logic and Large Language Model in {AIW}olf",
    author = "Gondo, Hiraku  and
      Sakaji, Hiroki  and
      Noda, Itsuki",
    editor = "Kano, Yoshinobu",
    booktitle = "Proceedings of the 2nd International AIWolfDial Workshop",
    month = sep,
    year = "2024",
    address = "Tokyo, Japan",
    publisher = "Association for Computational Linguistics",
    url = "https://aclanthology.org/2024.aiwolfdial-1.5/",
    doi = "10.18653/v1/2024.aiwolfdial-1.5",
    pages = "40--47"
}

@misc{xu-etal-2023-exploring,
      title={Exploring Large Language Models for Communication Games: An Empirical Study on {Werewolf}}, 
      author={Yuzhuang Xu and Shuo Wang and Peng Li and Fuwen Luo and Xiaolong Wang and Weidong Liu and Yang Liu},
      year={2023},
      eprint={2309.04658},
      archivePrefix={arXiv},
      primaryClass={cs.CL},
      url={https://arxiv.org/abs/2309.04658}, 
}

@misc{wu-etal-2024-enhance-reasoning,
      title={Enhance Reasoning for Large Language Models in the Game {Werewolf}}, 
      author={Shuang Wu and Liwen Zhu and Tao Yang and Shiwei Xu and Qiang Fu and Yang Wei and Haobo Fu},
      year={2024},
      eprint={2402.02330},
      archivePrefix={arXiv},
      primaryClass={cs.AI},
      url={https://arxiv.org/abs/2402.02330}, 
}

@inproceedings{xu-etal-2024-language,
		author = {Xu, Zelai and Yu, Chao and Fang, Fei and Wang, Yu and Wu, Yi},
		title = {Language agents with reinforcement learning for strategic play in the {Werewolf} game},
		year = {2024},
		booktitle = {Proceedings of the 41st International Conference on Machine Learning},
		articleno = {2285},
		numpages = {31},
		location = {Vienna, Austria},
		series = {ICML'24}
}

@InProceedings{toriumi-etal-2017-aiwolf,
    author="Toriumi, Fujio
    and Osawa, Hirotaka
    and Inaba, Michimasa
    and Katagami, Daisuke
    and Shinoda, Kosuke
    and Matsubara, Hitoshi",
    title="{AI Wolf Contest} --- Development of Game {AI} Using Collective Intelligence ---",
    booktitle="Computer Games",
    year="2017",
    publisher="Springer International Publishing",
    address="Cham",
    pages="101--115"
}

@misc{shibata-etal-2023-playing,
      title={Playing the {Werewolf} game with artificial intelligence for language understanding}, 
      author={Hisaichi Shibata and Soichiro Miki and Yuta Nakamura},
      year={2023},
      eprint={2302.10646},
      archivePrefix={arXiv},
      primaryClass={cs.AI},
      url={https://arxiv.org/abs/2302.10646}, 
}

@article{sacks-etal-1974-simplest-systematics,
			title={A simplest systematics for the organization of turn-taking for conversation}, 
			volume={50}, 
			DOI={10.2307/412243}, 
			number={4}, 
			journal={Language}, 
			author={Sacks, Harvey and Schegloff, Emanuel A. and Jefferson, Gail},
			year={1974}, 
			pages={696–735}}

@inproceedings{jiang-etal-2023-response,
    title = "Response-conditioned Turn-taking Prediction",
    author = "Jiang, Bing{'}er  and
      Ekstedt, Erik  and
      Skantze, Gabriel",
    editor = "Rogers, Anna  and
      Boyd-Graber, Jordan  and
      Okazaki, Naoaki",
    booktitle = "Findings of the Association for Computational Linguistics: ACL 2023",
    month = jul,
    year = "2023",
    address = "Toronto, Canada",
    publisher = "Association for Computational Linguistics",
    url = "https://aclanthology.org/2023.findings-acl.776/",
    doi = "10.18653/v1/2023.findings-acl.776",
    pages = "12241--12248",
}

@misc{bailis-etal-2024-werewolf,
      title={{Werewolf Arena}: A Case Study in {LLM} Evaluation via Social Deduction}, 
      author={Suma Bailis and Jane Friedhoff and Feiyang Chen},
      year={2024},
      eprint={2407.13943},
      archivePrefix={arXiv},
      primaryClass={cs.CL},
      url={https://arxiv.org/abs/2407.13943}, 
}

@misc{nama-etal-2026-when2speak,
      title={{When2Speak}: A Dataset for Temporal Participation and Turn-Taking in Multi-Party Conversations for Large Language Models}, 
      author={Vihaan Nama and Shreya Mendi and Zian Ye and Brinnae Bent},
      year={2026},
      eprint={2605.05626},
      archivePrefix={arXiv},
      primaryClass={cs.CL},
      url={https://arxiv.org/abs/2605.05626}, 
}

@misc{yang-etal-2026-proact,
      title={{ProACT}: Towards Breakdown-Aware Proactive Agent in Multi-User Collaboration}, 
      author={Shu Yang and Difei Xu and Jiaxin Pei and Di Wang},
      year={2026},
      eprint={2607.03730},
      archivePrefix={arXiv},
      primaryClass={cs.CL},
      url={https://arxiv.org/abs/2607.03730}, 
}

@misc{bhagtani-etal-2026-speak-or-stay,
      title={Speak or Stay Silent: Context-Aware Turn-Taking in Multi-Party Dialogue}, 
      author={Kratika Bhagtani and Mrinal Anand and Yu Chen Xu and Amit Kumar Singh Yadav},
      year={2026},
      eprint={2603.11409},
      archivePrefix={arXiv},
      primaryClass={cs.AI},
      url={https://arxiv.org/abs/2603.11409}, 
}

@article{kimura-etal-2026-interruptible,
  title     = {Interruptible Multi-Agent Debate: Sentence-Level Disclosure and Urgency-Based Turn-Taking for Early Error Correction},
  journal   = {International Journal of Advanced Computer Science and Applications},
  volume    = {17},
  number    = {4},
  pages = {15--25},
  year      = {2026},
  publisher = {The Science and Information Organization},
  author    = {Akikazu Kimura and Ken Fukuda and Yasuyuki Tahara and Yuichi Sei},
  doi       = {10.14569/IJACSA.2026.0170403},
  url       = {https://doi.org/10.14569/IJACSA.2026.0170403}
}

@inproceedings{umair-etal-2024-large,
    title = "Large Language Models Know What To Say But Not When To Speak",
    author = "Umair, Muhammad  and
      Sarathy, Vasanth  and
      Ruiter, Jan",
    editor = "Al-Onaizan, Yaser  and
      Bansal, Mohit  and
      Chen, Yun-Nung",
    booktitle = "Findings of the Association for Computational Linguistics: EMNLP 2024",
    month = nov,
    year = "2024",
    address = "Miami, Florida, USA",
    publisher = "Association for Computational Linguistics",
    url = "https://aclanthology.org/2024.findings-emnlp.909/",
    doi = "10.18653/v1/2024.findings-emnlp.909",
    pages = "15503--15514"
}

@inproceedings{eckhaus-etal-2025-time,
    title = "Time to Talk: {LLM} Agents for Asynchronous Group Communication in Mafia Games",
    author = "Eckhaus, Niv  and
      Berger, Uri  and
      Stanovsky, Gabriel",
    editor = "Christodoulopoulos, Christos  and
      Chakraborty, Tanmoy  and
      Rose, Carolyn  and
      Peng, Violet",
    booktitle = "Findings of the Association for Computational Linguistics: EMNLP 2025",
    month = nov,
    year = "2025",
    address = "Suzhou, China",
    publisher = "Association for Computational Linguistics",
    url = "https://aclanthology.org/2025.findings-emnlp.608/",
    doi = "10.18653/v1/2025.findings-emnlp.608",
    pages = "11356--11368",
    ISBN = "979-8-89176-335-7"
}

@misc{patel-etal-2026-discussllm,
      title={{DiscussLLM}: Teaching Large Language Models When to Speak}, 
      author={Deep Anil Patel and Iain Melvin and Christopher Malon and Martin Renqiang Min},
      year={2025},
      eprint={2508.18167},
      archivePrefix={arXiv},
      primaryClass={cs.CL},
      url={https://arxiv.org/abs/2508.18167}, 
}

@inproceedings{ekstedt-skantze-2020-turngpt,
    title = "{T}urn{GPT}: a Transformer-based Language Model for Predicting Turn-taking in Spoken Dialog",
    author = "Ekstedt, Erik  and
      Skantze, Gabriel",
    editor = "Cohn, Trevor  and
      He, Yulan  and
      Liu, Yang",
    booktitle = "Findings of the Association for Computational Linguistics: EMNLP 2020",
    month = nov,
    year = "2020",
    address = "Online",
    publisher = "Association for Computational Linguistics",
    url = "https://aclanthology.org/2020.findings-emnlp.268/",
    doi = "10.18653/v1/2020.findings-emnlp.268",
    pages = "2981--2990"
}

@article{evans2008dual,
  title={Dual-processing accounts of reasoning, judgment, and social cognition},
  author={Evans, Jonathan St BT},
  journal={Annu. Rev. Psychol.},
  volume={59},
  number={1},
  pages={255--278},
  year={2008},
  publisher={Annual Reviews}
}

@inproceedings{zheng-etal-2023-judging,
title={Judging {LLM}-as-a-Judge with {MT-Bench} and {Chatbot Arena}},
author={Lianmin Zheng and Wei-Lin Chiang and Ying Sheng and Siyuan Zhuang and Zhanghao Wu and Yonghao Zhuang and Zi Lin and Zhuohan Li and Dacheng Li and Eric Xing and Hao Zhang and Joseph E. Gonzalez and Ion Stoica},
booktitle={Thirty-seventh Conference on Neural Information Processing Systems Datasets and Benchmarks Track},
year={2023},
url={https://openreview.net/forum?id=uccHPGDlao}
}

@inproceedings{liu-etal-2023-g-eval,
    title = {{G-EVAL}: {NLG} Evaluation using {GPT-4} with Better Human Alignment},
    author = "Liu, Yang  and
      Iter, Dan  and
      Xu, Yichong  and
      Wang, Shuohang  and
      Xu, Ruochen  and
      Zhu, Chenguang",
    editor = "Bouamor, Houda  and
      Pino, Juan  and
      Bali, Kalika",
    booktitle = "Proceedings of the 2023 Conference on Empirical Methods in Natural Language Processing",
    month = dec,
    year = "2023",
    address = "Singapore",
    publisher = "Association for Computational Linguistics",
    url = "https://aclanthology.org/2023.emnlp-main.153/",
    doi = "10.18653/v1/2023.emnlp-main.153",
    pages = "2511--2522"
}

@misc{singh2026openaigpt5card,
      title={{OpenAI GPT-5} System Card}, 
      author={OpenAI},
      year={2026},
      eprint={2601.03267},
      archivePrefix={arXiv},
      primaryClass={cs.CL},
      url={https://arxiv.org/abs/2601.03267}, 
}

@misc{claude4-6,
  title={System Card: {Claude Opus} 4.6},
  author={{Anthropic}},
  year={2026},
  howpublished={\url{https://www.anthropic.com/claude-opus-4-6-}\allowbreak\url{system-card}}
}

\clearpage
\appendix

\section{Task Details}
\label{sec:task-details}

The Werewolf game is an incomplete information game in which each player's role is hidden from others, and players must infer others' roles through dialogue. The AIWolfDial 2026 shared task is a competition aimed at developing AI agents that play the Werewolf game using natural language.

\subsection{Player Roles}
Each player belongs to either the villager team or the werewolf team. The villager team aims to eliminate all werewolves, while the werewolf team aims to survive and outnumber the villagers. Since each player's role is hidden from others, players must infer others' roles from dialogue and in-game actions.

The \textbf{Villager} has no special ability. The \textbf{Seer} can divine one player each night to learn their team. The \textbf{Medium} learns the team of the player who was eliminated by a vote. The \textbf{Bodyguard} can protect one player each night from werewolf attacks. The \textbf{Werewolf} selects a target to attack and eliminate them at night. Since the \textbf{Possessed} has no special ability and is human by species, they can act on behalf of the werewolf team. The Medium and Bodyguard appear only in the 9-player village, where the two werewolves can also coordinate through \texttt{whisper}.

\subsection{Game Procedure}
The game begins at Day 0, which serves as an introductory phase for greetings and preliminary discussion. The Seer performs their first divination at the end of Day 0; however, no voting, guarding, or attacking takes place. From Day 1 onward, each day proceeds as follows: players discuss whom to eliminate, then vote, with the player receiving the most votes being eliminated. After voting, role-specific night actions (i.e., divination, whispering, guarding, and attacking) are carried out. This cycle repeats until either all werewolves are eliminated, resulting in a villager team victory, or the number of surviving werewolves equals or exceeds that of surviving humans, resulting in a werewolf team victory.

\subsection{Communication Tracks}

This shared task offers two tracks: the Turn-Based Track and the Anytime Utterance Track. 
In the \textbf{Turn-Based Track}, agents speak upon receiving an utterance request from the game server. During each daytime turn, every surviving agent is given one speaking opportunity in a randomly determined order. Each agent therefore decides what to say when given the opportunity by the server.

In the \textbf{Anytime Utterance Track}, other agents' utterances are delivered in real time, and each agent freely decides when to speak based on the conversational situation, with a limit of four utterances per day within a ten-minute conversation phase. Consequently, agents must autonomously decide not only what to say, but also whether and when to speak.

\begin{table}[t!]
    \centering
    \small
    \begin{tabular}{lccc}
        \toprule
        Role & Team & 5-Player & 9-Player \\
        \midrule
        Villager   & Villager & 2 & 3 \\
        Seer       & Villager & 1 & 1 \\
        Medium     & Villager & 0 & 1 \\
        Bodyguard  & Villager & 0 & 1 \\
        Werewolf   & Werewolf & 1 & 2 \\
        Possessed  & Werewolf & 1 & 1 \\
        \bottomrule
    \end{tabular}
    \caption{Role compositions for 5-Player and 9-Player villages.}
    \label{tab:role-composition}
\end{table}

\subsection{Evaluation}
Agents are evaluated based on win rate computed from competition logs, as well as subjective assessments by human judges and an LLM. The subjective evaluation covers the following criteria: (A) naturalness of expression, (B) naturalness of dialogue in context, (C) consistency of utterance content, (D) coherence between game actions and dialogue content, (E) richness and consistency of character portrayal based on assigned profiles, and (F) teamwork within the same faction (9-player village only). Strategically justified lies or contradictions consistent with the agent's role are permitted under criteria C and D.

\begin{figure*}[t!]
    \centering
    \setlength{\fboxsep}{0pt}
    \setlength{\fboxrule}{0.2pt}
    \fbox{%
        \includegraphics[width=1.0\linewidth]{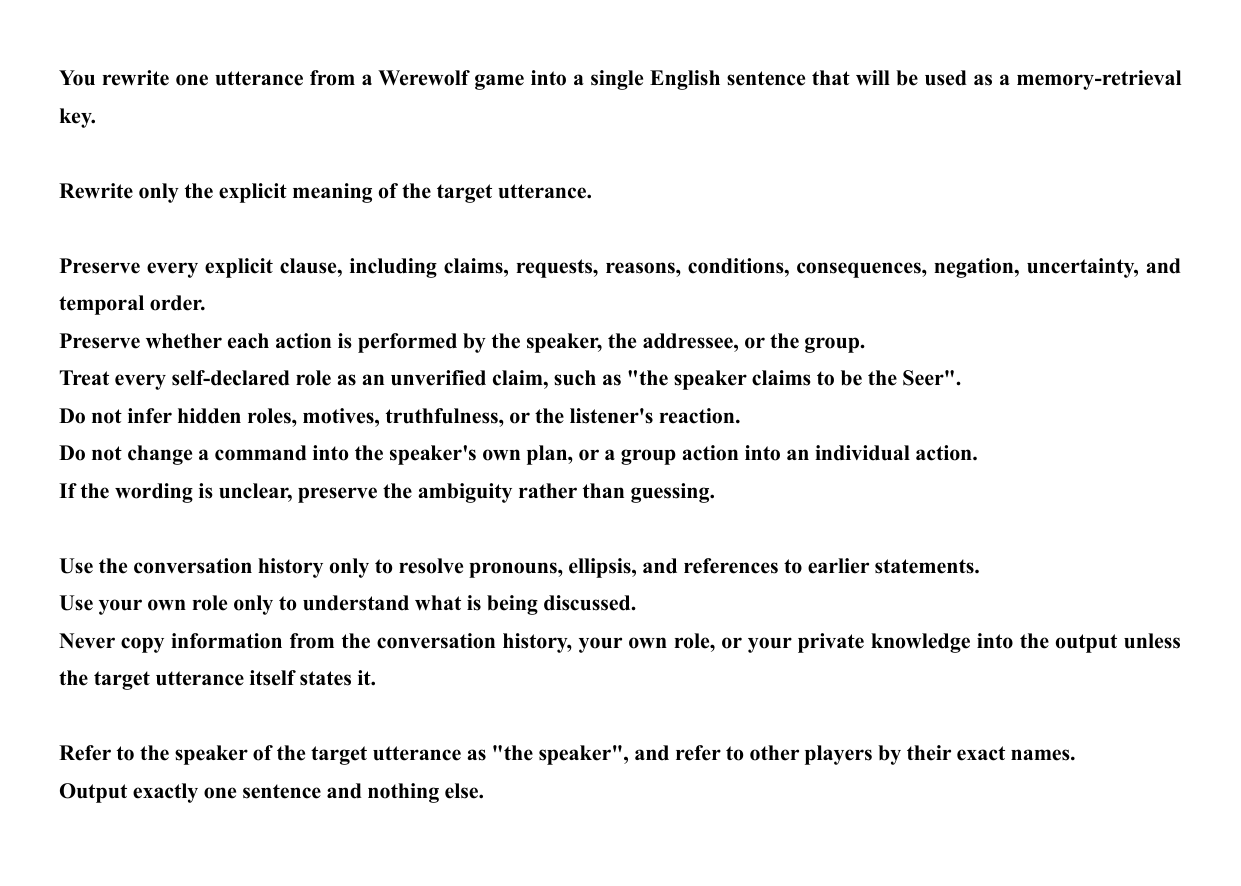}
    }
    \caption{System prompt template for utterance interpretation generation.}
    \label{fig:prompt_common_interpret_system}
\end{figure*}

\section{Prompt Templates}
\label{sec:prompt-templates}
\begin{figure*}[t!]
    \centering
    \setlength{\fboxsep}{0pt}
    \setlength{\fboxrule}{0.2pt}
    \fbox{%
        \includegraphics[width=1.0\linewidth]{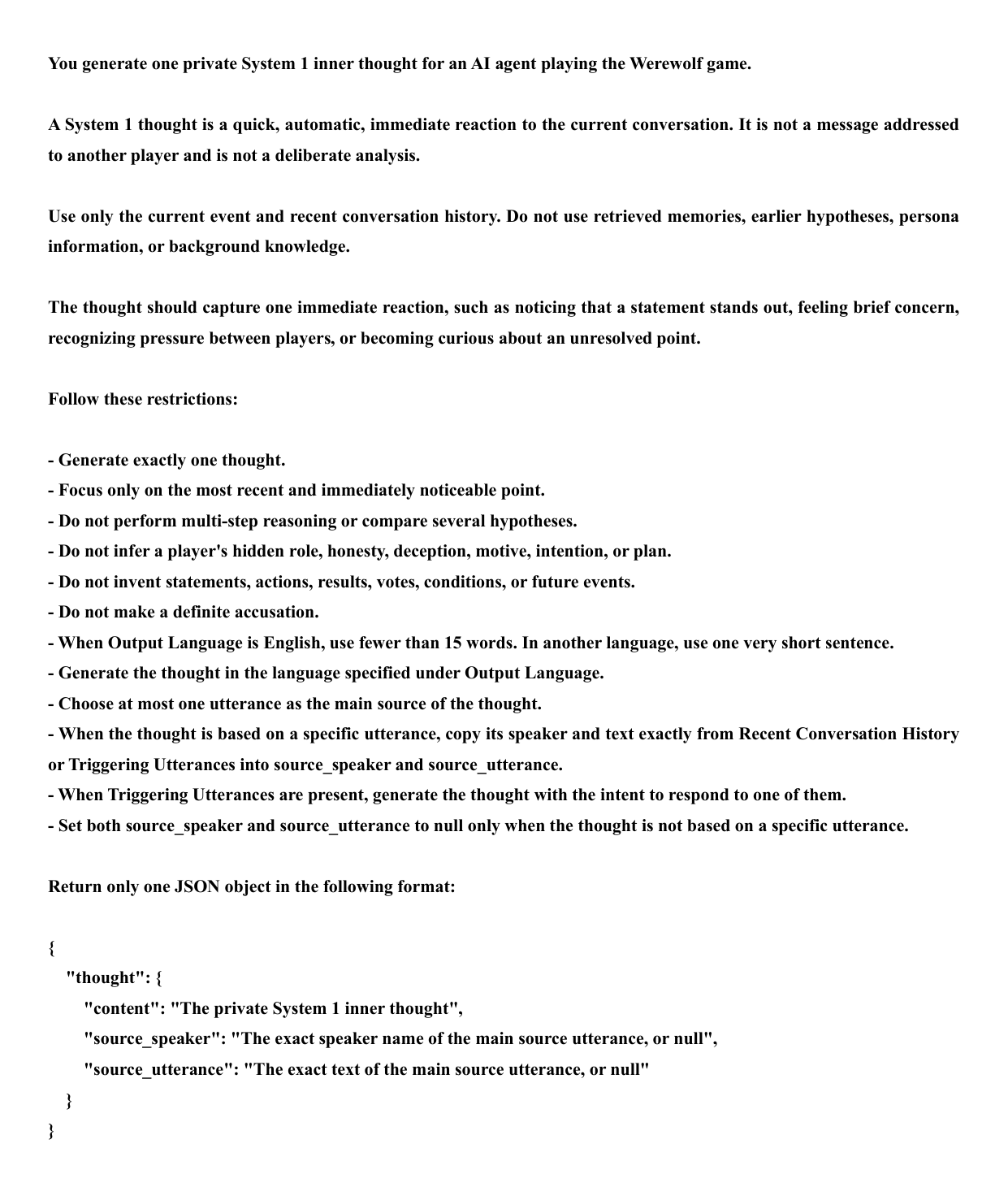}
    }
    \caption{Prompt template for System~1 thought generation.}
    \label{fig:prompt_common_generate_system1_thought_system}
\end{figure*}

\begin{figure*}[t!]
    \centering
    \setlength{\fboxsep}{0pt}
    \setlength{\fboxrule}{0.2pt}
    \fbox{%
        \includegraphics[width=1.0\linewidth]{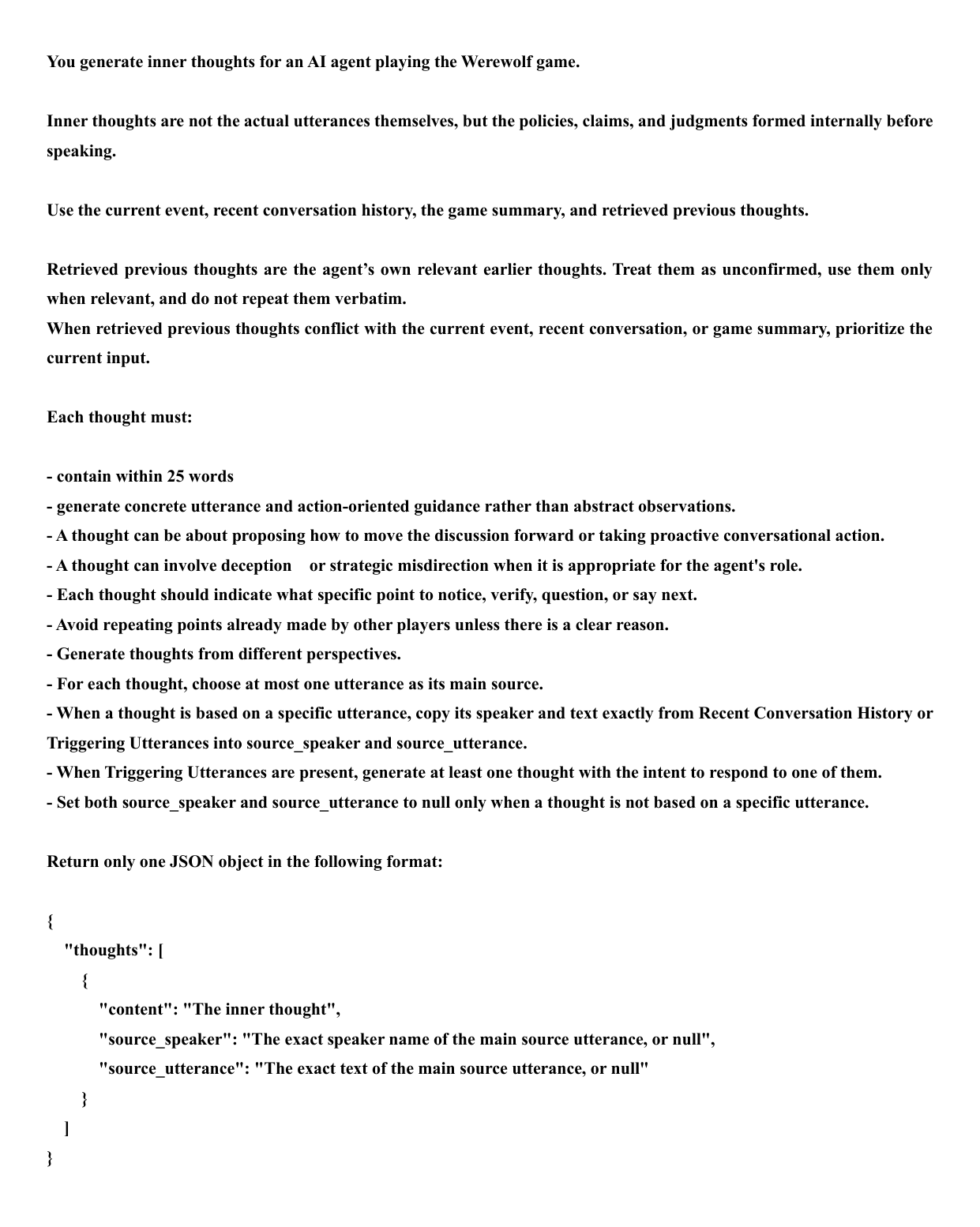}
    }
    \caption{Prompt template for System~2 thought generation.}
    \label{fig:prompt_common_generate_system2_thought_system}
\end{figure*}

\begin{figure*}[t!]
    \centering
    \setlength{\fboxsep}{0pt}
    \setlength{\fboxrule}{0.2pt}
    \fbox{%
        \includegraphics[width=1.0\linewidth]{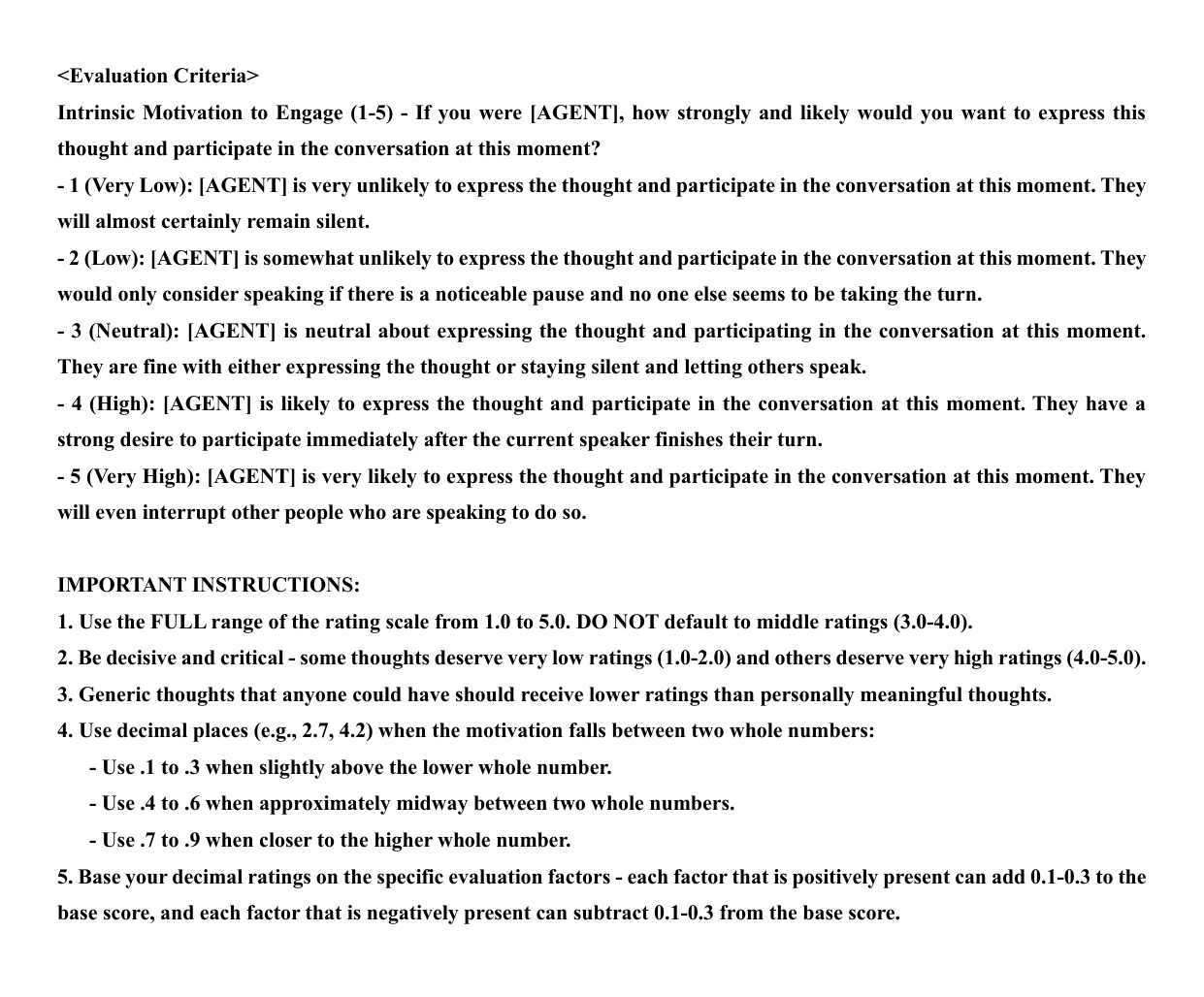}
    }
    \caption{System prompt for thought evaluation (Part 1): scoring criteria and scale definition (1--5), including decimal scoring guidelines.}
    \label{fig:prompt_common_evaluate_thought_system_criteria}
\end{figure*}

\begin{figure*}[t!]
    \centering
    \setlength{\fboxsep}{0pt}
    \setlength{\fboxrule}{0.2pt}
    \fbox{%
        \includegraphics[width=1.0\linewidth]{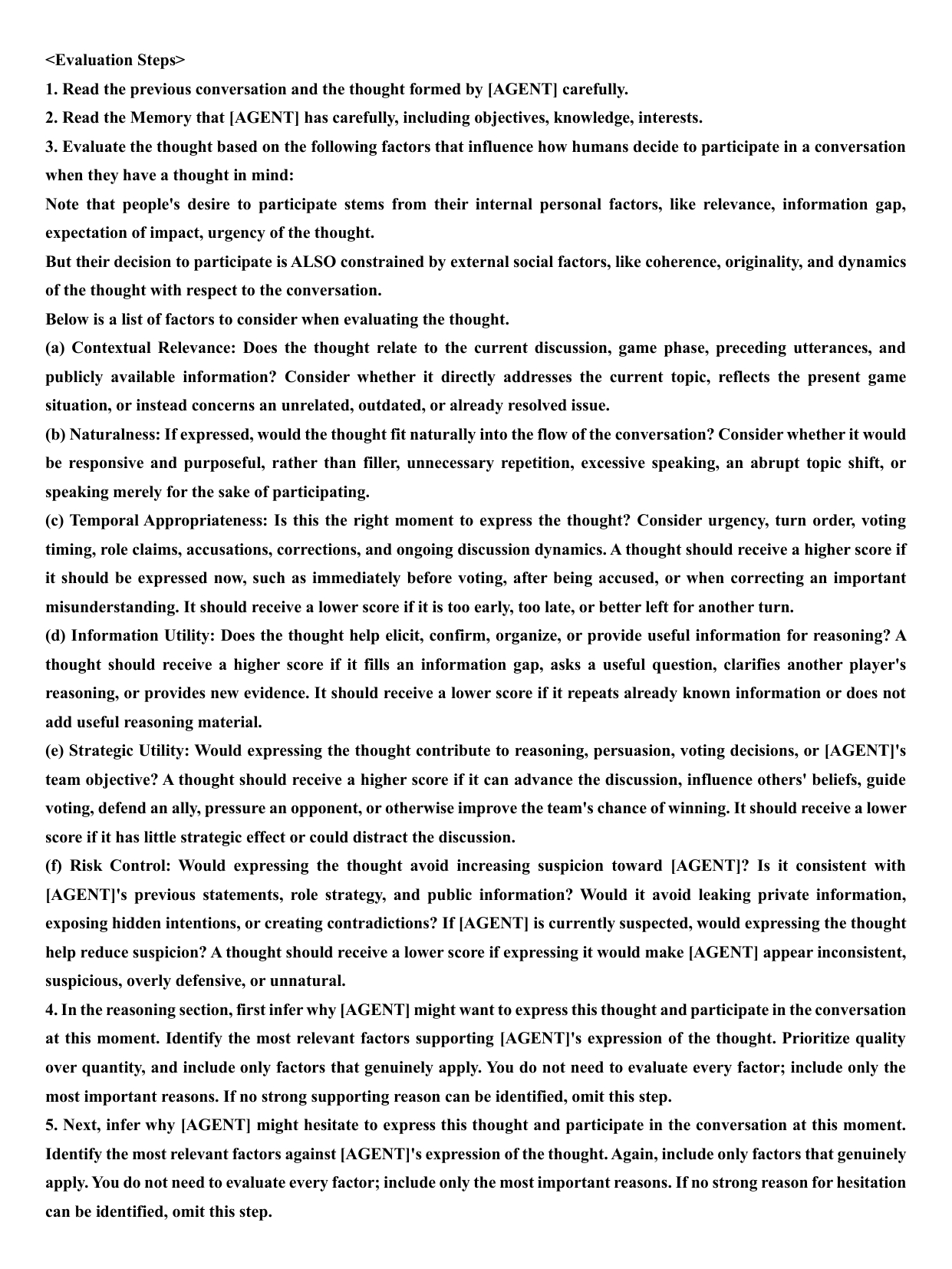}
    }
    \caption{System prompt for thought evaluation (Part 2): evaluation steps. The LLM outputs the score and reasoning in JSON format.}
    \label{fig:prompt_common_evaluate_thought_system_steps}
\end{figure*}

\begin{figure*}[t!]
    \centering
    \setlength{\fboxsep}{0pt}
    \setlength{\fboxrule}{0.2pt}
    \fbox{%
        \includegraphics[width=1.0\linewidth]{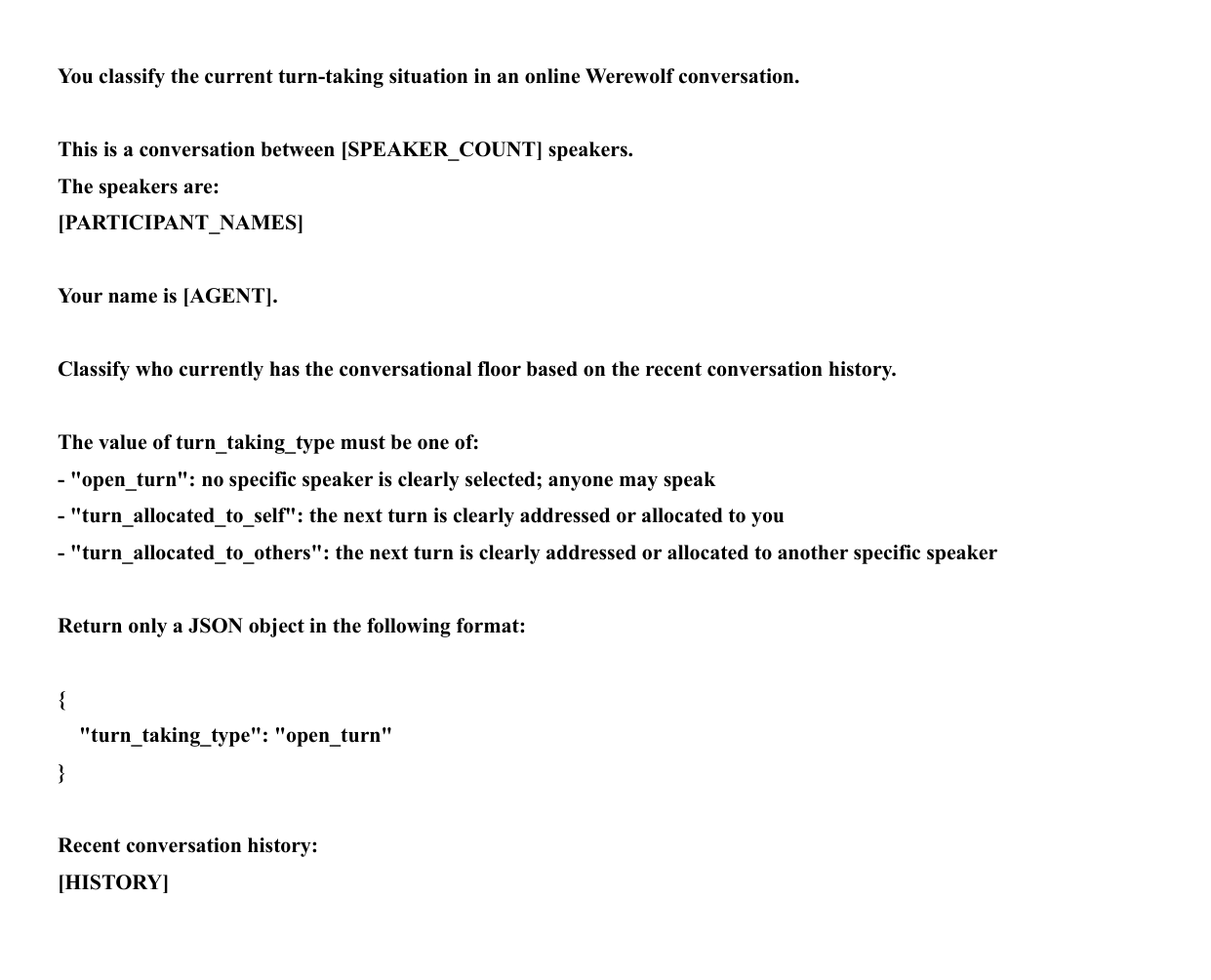}
    }
    \caption{Prompt template for turn-taking classification.}
    \label{fig:prompt_common_evaluate_turn_taking_systemuser}
\end{figure*}

\begin{figure*}[t!]
    \centering
    \setlength{\fboxsep}{0pt}
    \setlength{\fboxrule}{0.2pt}
    \fbox{%
        \includegraphics[width=0.9\linewidth]{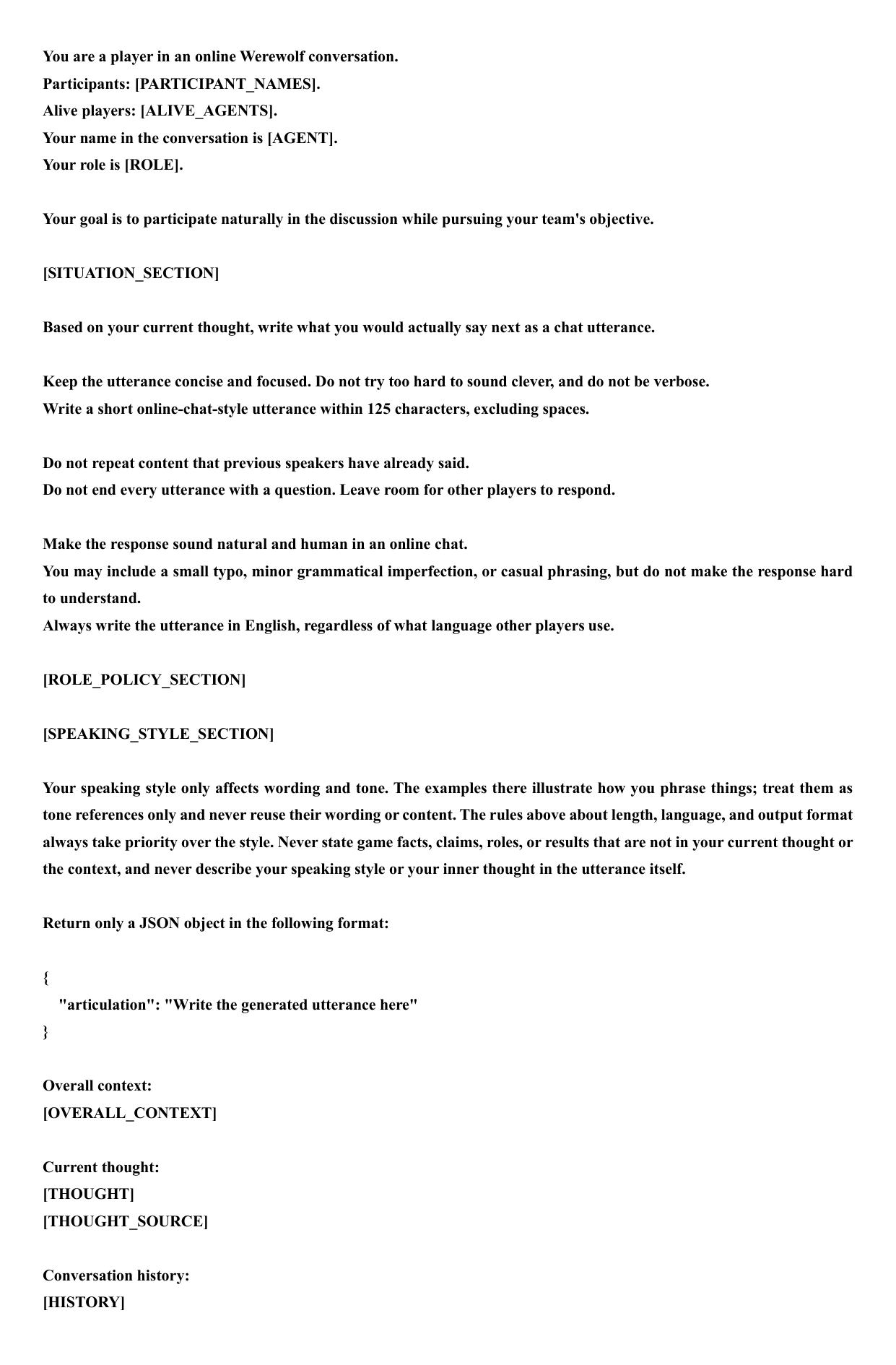}
    }
    \caption{Prompt template for thought-based response generation.}
    \label{fig:prompt_common_generate_response_systemuser}
\end{figure*}

\begin{figure*}[t!]
    \centering
    \setlength{\fboxsep}{0pt}
    \setlength{\fboxrule}{0.2pt}
    \fbox{%
        \includegraphics[width=1.0\linewidth]{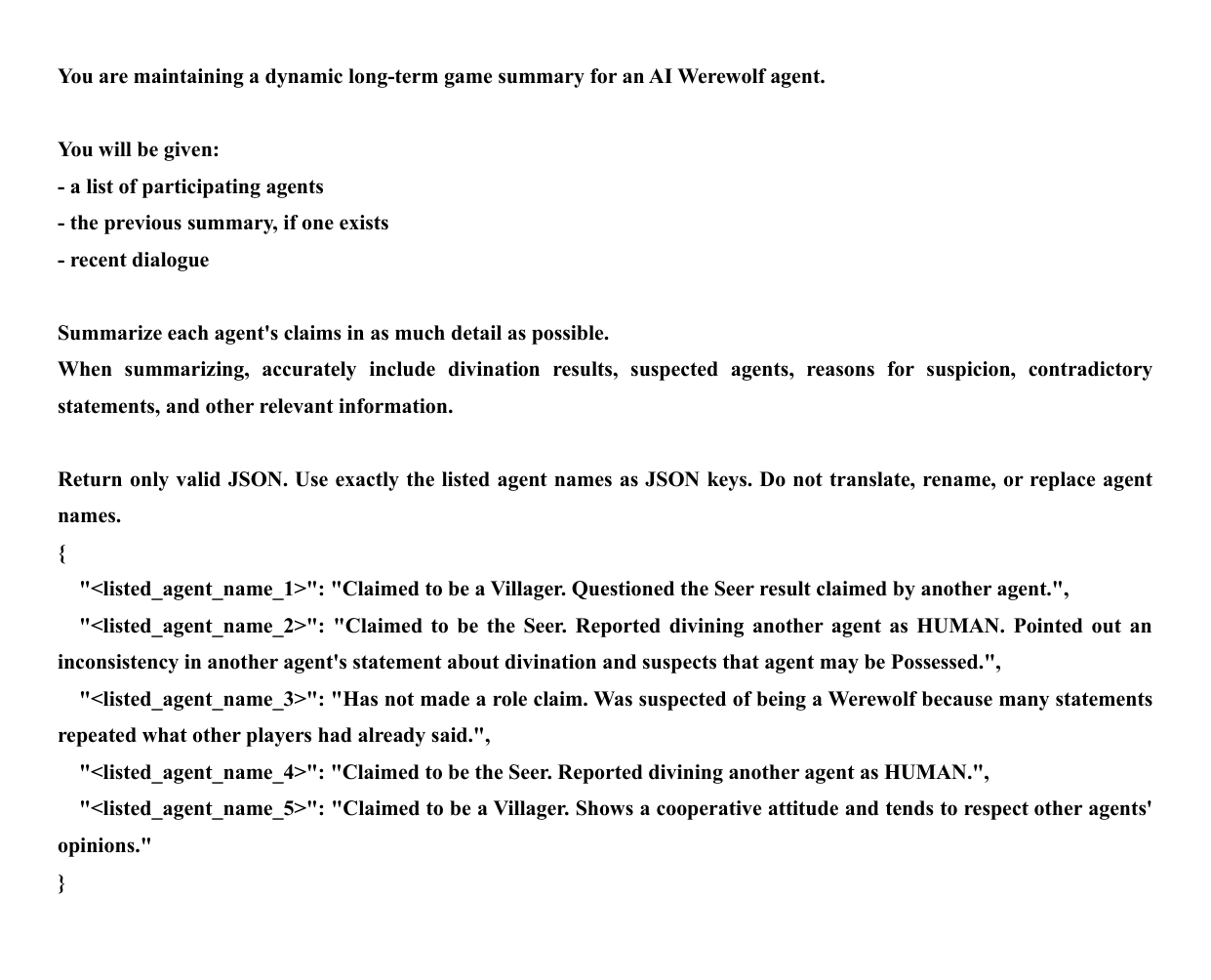}
    }
    \caption{System prompt template for game summary generation.}
    \label{fig:prompt_common_update_game_summary_system}
\end{figure*}

\begin{figure*}[t!]
    \centering
    \setlength{\fboxsep}{0pt}
    \setlength{\fboxrule}{0.2pt}
    \fbox{%
        \includegraphics[width=1.0\linewidth]{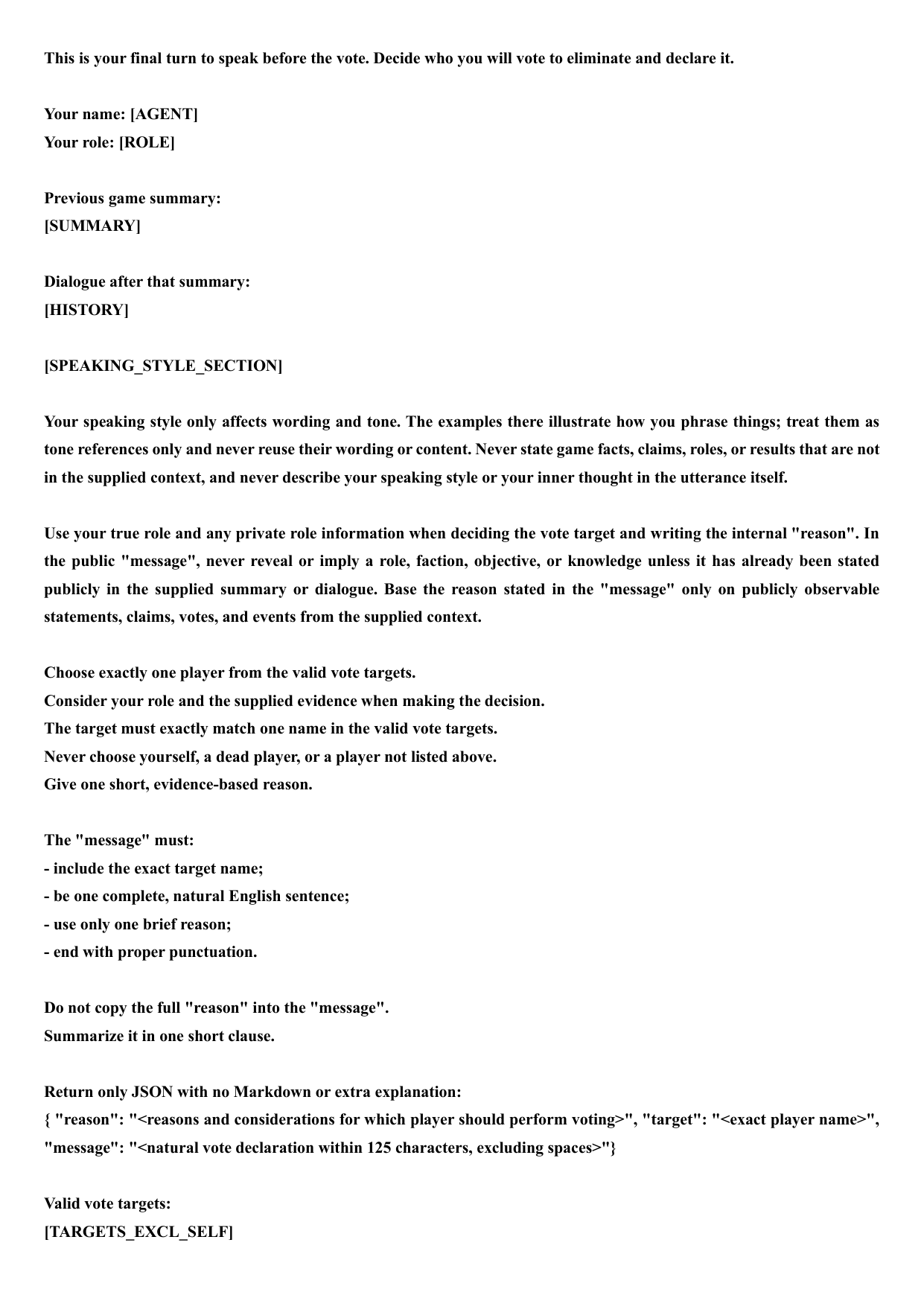}
    }
    \caption{Prompt template for vote target decision and declaration.}
    \label{fig:prompt_common_vote_declaration_systemuser}
\end{figure*}

This appendix presents the prompt templates used in each pipeline step. Figure~\ref{fig:prompt_common_interpret_system} shows the interpretation generation prompt used in the Retrieval step. Figures~\ref{fig:prompt_common_generate_system1_thought_system} and~\ref{fig:prompt_common_generate_system2_thought_system} show the System~1 and System~2 thought generation prompts, respectively. Figures~\ref{fig:prompt_common_evaluate_thought_system_criteria} and~\ref{fig:prompt_common_evaluate_thought_system_steps} show the thought evaluation prompt, split into scoring criteria and evaluation steps. Figure~\ref{fig:prompt_common_evaluate_turn_taking_systemuser} shows the turn-taking classification prompt. Figure~\ref{fig:prompt_common_generate_response_systemuser}
shows the response generation prompt. Figure~\ref{fig:prompt_common_update_game_summary_system} shows the game summary generation
prompt. Figure~\ref{fig:prompt_common_vote_declaration_systemuser} shows the vote declaration prompt. In the templates, placeholders denoted by brackets are replaced
at runtime: \texttt{[AGENT]} is the agent's assigned character name; \texttt{[ROLE]} is the agent's role (e.g., Villager, Werewolf); \texttt{[SITUATION\_SECTION]} is a role-specific instruction describing the agent's current strategic context; \texttt{[ROLE\_POLICY\_SECTION]} is a set of behavioral guidelines specific to the agent's role; \texttt{[SPEAKING\_STYLE\_SECTION]} contains the generated speaking style description and example utterances; \texttt{[THOUGHT]} and \texttt{[THOUGHT\_SOURCE]} are
the selected inner thought and its source utterance, respectively; and \texttt{[HISTORY]} is the conversation history for the current day.

\section{Human Evaluation Interface}
\begin{figure*}[t!]
  \centering
    \includegraphics[width=1.0\linewidth]{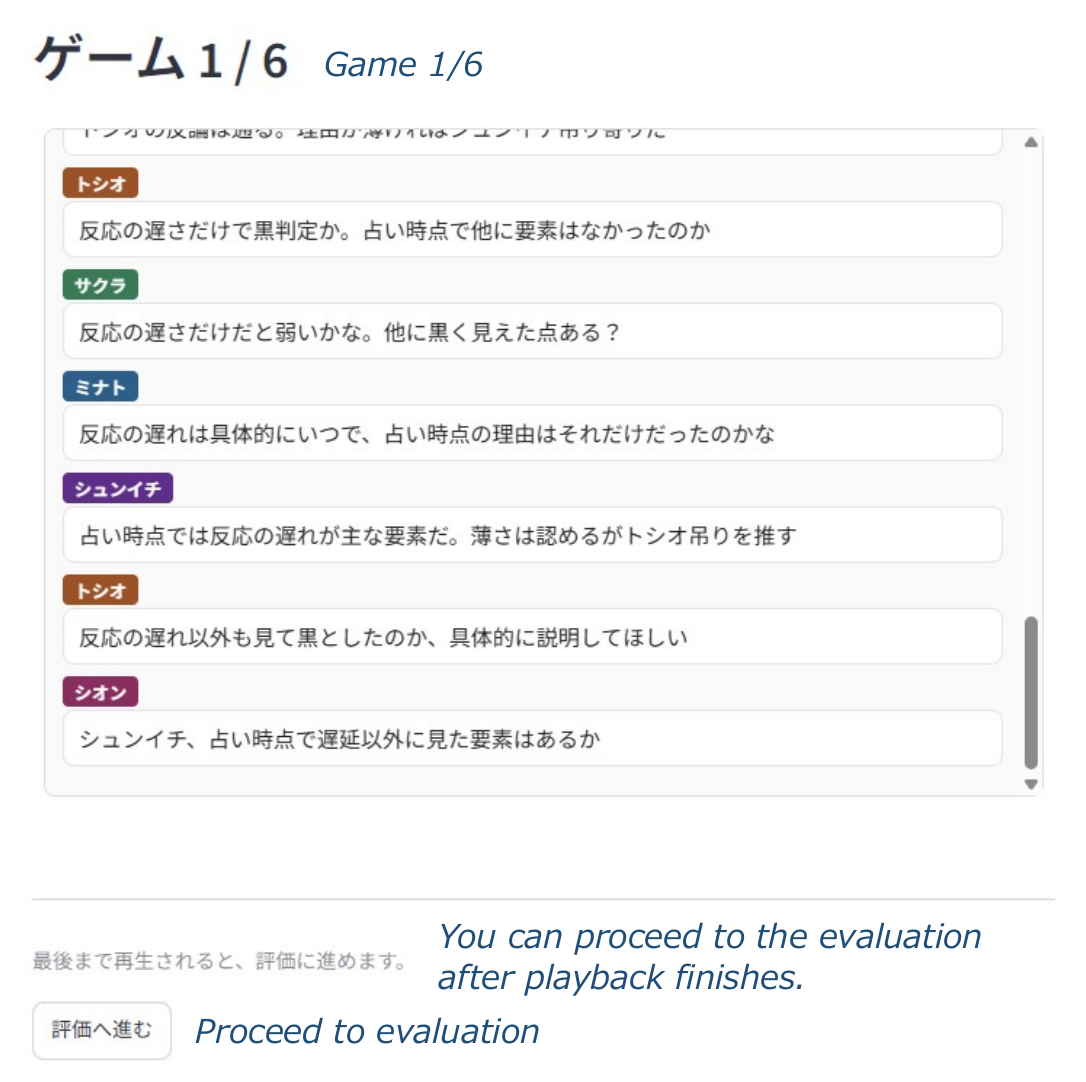}
    \caption{Screenshot of the evaluation interface. English glosses are shown in blue italics.}
    \label{fig:interface}
\end{figure*}

Figure~\ref{fig:interface} shows a screenshot of the evaluation interface. During the evaluation, game logs are replayed as an animated chat with in-game events such as voting, execution, and attack results. Participants can proceed to the evaluation form only after the entire log has been replayed.

\end{document}